\documentclass[A4,12pt]{article}

\usepackage{amssymb}
\usepackage{amsmath}
\usepackage{amsthm}

\usepackage{subfig}

  \usepackage{algorithm}
  \usepackage{algpseudocode}
 \newcommand*\Let[2]{\State #1 $\gets$ #2}

\newcommand{\R}{\mathbb{R}}
\DeclareMathOperator*{\argmin}{arg\,min}

\usepackage{todonotes}

\newcommand{\GroveTwoSSIM}{0.6920}
\newcommand{\GroveTwoLTwo}{3.70e-05}
\newcommand{\GroveTwoSNR}{72.9669}
\newcommand{\GroveTwoEPE}{1.0290}
\newcommand{\GroveTwoAE}{0.1713}

\newcommand{\GroveThreeSSIM}{0.6036}
\newcommand{\GroveThreeLTwo}{5.62e-05}
\newcommand{\GroveThreeSNR}{69.2809}
\newcommand{\GroveThreeEPE}{1.8428}
\newcommand{\GroveThreeAE}{0.2115}

\newcommand{\HydrangeaSSIM}{0.7412}
\newcommand{\HydrangeaLTwo}{3.85e-05}
\newcommand{\HydrangeaSNR}{74.0326}
\newcommand{\HydrangeaEPE}{1.3262}
\newcommand{\HydrangeaAE}{0.1396}

\newcommand{\RubberWhaleSSIM}{0.8350}
\newcommand{\RubberWhaleLTwo}{2.10e-05}
\newcommand{\RubberWhaleSNR}{79.2496}
\newcommand{\RubberWhaleEPE}{0.6566}
\newcommand{\RubberWhaleAE}{0.2514}

\newcommand{\UrbanTwoSSIM}{0.7103}
\newcommand{\UrbanTwoLTwo}{2.91e-05}
\newcommand{\UrbanTwoSNR}{75.1465}
\newcommand{\UrbanTwoEPE}{2.0765}
\newcommand{\UrbanTwoAE}{0.1588}

\newcommand{\UrbanThreeSSIM}{0.6961}
\newcommand{\UrbanThreeLTwo}{3.11e-05}
\newcommand{\UrbanThreeSNR}{74.4912}
\newcommand{\UrbanThreeEPE}{2.2277}
\newcommand{\UrbanThreeAE}{0.2395}

\title{Joint Large-Scale Motion Estimation and Image Reconstruction}

\author{Hendrik Dirks\thanks{Institute for Computational and Applied Mathematics and Cells in Motion Cluster of Excellence, University of M{\"u}nster, Orl\'{e}ans-Ring 10, 48149 M{\"u}nster, Germany, Email: hendrik.dirks@wwu.de}}

\begin{document}
	
	\maketitle
	

\begin{abstract}
This article describes the implementation of the joint motion estimation and image reconstruction framework presented by Burger, Dirks and Sch\"onlieb and extends this framework to large-scale motion between consecutive image frames.\\
The variational framework uses displacements between consecutive frames based on the optical flow approach to improve the image reconstruction quality on the one hand and the motion estimation quality on the other. The energy functional consists of a data-fidelity term with a general operator that connects the input sequence to the solution, it has a total variation term for the image sequence and is connected to the underlying flow using an optical flow term. Additional spatial regularity for the flow is modeled by a total variation regularizer for both components of the flow. \\
The numerical minimization is performed in an alternating manner using primal-dual techniques. The resulting schemes are presented as pseudo-code together with a short numerical evaluation.
\end{abstract}


\section{Introduction}
Image reconstruction and motion estimation are important tasks in image processing. Such problems arise for example in modern medicine, biology, chemistry or physics, where even the smallest objects are observed by high resolution microscopes. To characterize the dynamics involved in such data, velocity fields between consecutive image frames are calculated. This is challenging, since the recorded images often suffer from low resolution, low contrast, different gray levels and noise. Methods that simultaneously denoise the recorded image sequence and calculate the underlying velocity field offer new opportunities, since both tasks may endorse each other. \\
The ansatz from \cite{dirks} aims at reconstructing a given sequence $u$ of images and calculating flow fields $\boldsymbol{v}$ between subsequent images at the same time. For given measurements $f=Ku$ this can be achieved by minimizing the variational model
\begin{align}
\label{equation:generalModelEQ}
&\int_0^T \frac{1}{2}\left\|K_tu(\cdot,t)-f(\cdot,t)\right\|_2^2 + \alpha \mathcal{R}(u(\cdot,t))  + \beta \mathcal{S}(\boldsymbol{v}(\cdot,t)) + \|\rho(u,\boldsymbol{v})\|_1 dt
\end{align}
with respect to $u$ and $\boldsymbol{v}$ simultaneously. The denoising part is based on the ROF model \cite{rudin1992nonlinear}. The first part $\left\|Ku-f\right\|_2^2$ connects the input data $f$ with the image sequence $u$ via a linear operator $K$. Depending on the application, $K$ may model the cutting out of a subset $\Sigma\subset\Omega$ for inpainting, a subsampling for super resolution, a blur for deconvolution or a Radon transform for computed tomography. Additional a-priori information about the structure of $u$ respectively $\boldsymbol{v}$ can be incorporated into each frame via the regularization terms $\mathcal{R}(u(\cdot,t))$ and $\mathcal{S}(\boldsymbol{v}(\cdot,t))$, while their significance is weighted using $\alpha$ and $\beta$. Finally, flow field and images are coupled by a model for the underlying dynamics, e.g. the optical flow approach $\rho(u,\boldsymbol{v}) = u_t+\nabla u\cdot\boldsymbol{v}$ (see Section \ref{subsection:noiseSensitivity}).\\
\subsection{Notations}
Let us for the following assume a discrete rectangular domain $\Omega\subset\R^2$ and sequences of recorded (noisy) images $f^1,\ldots,f^n : \Omega\rightarrow\R$, clean images $u^1,\ldots,u^n : \Omega\rightarrow\R$ and motion fields between subsequent frames $\boldsymbol{v}^1,\ldots,\boldsymbol{v}^{n-1} : \Omega\rightarrow\R^2$.

\subsection{Optical Flow}
\label{subsection:noiseSensitivity}
In the optical flow problem we want to estimate the flow field $\boldsymbol{v}^i = (v^{i,1},v^{i,2})^T$ that describes the displacement between each of the subsequent images $u^{i}$ and $u^{i+1}$. For this sake it is common to use the brightness constancy assumption 
\begin{equation}
u^{i+1}(x+\boldsymbol{v}^i(x)) - u^i(x) = 0,\quad i=1,\ldots,n-1,x\in\Omega
\label{opticalFlowKomplett}
\end{equation}
to derive a connection between image intensities and the underlying flow. Due to the non-linearity (in terms of $\boldsymbol{v}^i(x)$) of this formulation, one often linearizes the first term and arrives at
\begin{equation}
\boldsymbol{v}^i\cdot\nabla u^{i+1}(x) + u^{i+1}(x) - u^i(x) = 0.
\label{opticalFlowKlassisch}
\end{equation}
Unfortunately, due to the Taylor expansion, equation \eqref{opticalFlowKlassisch} is only valid for small displacements. Another way, which in theory may handle displacements of arbitrary magnitude, is to use a given flow field $\boldsymbol{\tilde{v}}^i$ and use a Taylor expansion of $u^{i+1}(x+\boldsymbol{v}^i(x))$ with respect to a known flow field $\boldsymbol{\tilde{v}}^i$. The result is again a linear equation for $\boldsymbol{v}^i$ which requires evaluations of the input images at a shifted domain $x+\boldsymbol{\tilde{v}^i}$
\begin{equation}
(\boldsymbol{v}^i-\boldsymbol{\tilde{v}}^i)\cdot\nabla u^{i+1}(x+\boldsymbol{\tilde{v}}^i) + u^{i+1}(x+\boldsymbol{\tilde{v}}^i) - u^i(x) = 0.
\label{opticalFlowBesser}
\end{equation}
By defining $\tilde{u}^{i+1} := u^{i+1}(x+\boldsymbol{\tilde{v}}^i)$ the left side of equation \eqref{opticalFlowBesser} becomes
\begin{equation}
\rho(\boldsymbol{v}^i,u^{i},u^{i+1}) := (\boldsymbol{v}^i-\boldsymbol{\tilde{v}}^i)\cdot  \nabla\tilde{u}^{i+1} + \tilde{u}^{i+1} - u^i.
\label{opticalFlowBesser2}
\end{equation}
Both, formulation \eqref{opticalFlowKlassisch} and \eqref{opticalFlowBesser}, state only one equation per point for the two unknown components of $\boldsymbol{v}$ and, consequently, the problem is underdetermined. To overcome this, the optical flow formulation can be used as a data fidelity in a variational model together with an isotropic total variation term on each of the two flow components to ensure spatial regularity. This leads to the following variational problem for calculating the optical flow of a video sequence $u^1,\ldots,u^n$:
\begin{equation}
	\argmin_{\boldsymbol{v} = \boldsymbol{v}^1,\ldots, \boldsymbol{v}^{n-1}} \sum_{i=1}^{n-1} \| \rho(\boldsymbol{v}^i,u^{i},u^{i+1}) \|_1 + \alpha \sum_{j=1}^2\left\|\nabla v^{i,j}\right\|_{1,2}.
	\label{variationalMotionModelGeneral}
\end{equation}
This variational approach to optical flow was first introduced by Horn and Schunck \cite{horn1981determining} and continuously improved \cite{aubert1999computing,papenberg2006highly}. Further details about the so-called L$^1$-TV optical flow model above can be found e.g. in \cite{wedel2009improved,zach2007duality}.

\subsection{Image Reconstruction}
Nowadays, variational models for image reconstruction have become very popular. One of the most famous models, introduced by Rudin, Osher and Fatemi in 1992 \cite{rudin1992nonlinear}, is the total variation (TV) model, where the authors couple a L$^2$ data fidelity term with a total variation regularization. Data-term and regularizer in the ROF model match with the first two terms model \eqref{equation:generalModelEQ}. The TV-regularization results in a denoised image with cartoon-like features. This model has also been adapted to image deblurring \cite{Wang07afast}, inpainting \cite{shen2002mathematical}, super resolution \cite{mitzel2009video,unger2010convex} and tomographic reconstruction \cite{sawatzky2008accurate,kosters2011emrecon}. We collectively call these image reconstruction models.\\
From our point of view, the problem of motion estimation is directly connected to the underlying image sequence and, hence, requires accurate input images $u^i$. Unfortunately, for many practical applications, only noisy variants $f^i$ of $u^i$ can be recorded. Similar to the optical flow problem, a variational formulation can be used to reconstruct $u^i$ from $f^i$. Let us assume that $f^i$ is a degraded version of $u^i$ corrupted by Gaussian noise. Then, the ROF model
\begin{equation}
	\argmin_{u = u^1,\ldots, u^{n}} \sum_{i=1}^{n} \frac{1}{2}\|A^iu^i-f^i \|_2^2 + \alpha \|\nabla u^i\|_{1,2}
	\label{variationalDenoisingModelGeneral}
\end{equation}
can be used to reconstruct each $u^i$ from $f^i$. Here, $A^i$ represents some linear operator and could be the identity (denoising), subsampling (zooming) or the Radon transform (computed tomography).
\subsection{Joint Model}
As already mentioned, motion estimation should be done on noise-free images, so generally one first denoises the image sequence using \eqref{variationalDenoisingModelGeneral} and afterwards estimates the underlying velocity fields using \eqref{variationalMotionModelGeneral}. In \cite{dirks} it has been shown that a joint model that simultaneously recovers an image sequence and estimates motion offers a significant advantage towards subsequently applying both methods. The following joint model
\begin{equation}
	\argmin_{u,\boldsymbol{v}} \int_0^T \frac{1}{2}\left\|Au-f\right\|_2^2 + \alpha \left\|\nabla u\right\|_{1,2} + \gamma \left\|u_t+\nabla u\cdot\boldsymbol{v}\right\|_1  + \beta\sum_{j=1}^2\left\|\nabla v_j\right\|_{1,2} dt,
	\label{oldJointModel}
\end{equation}
respectively its time-discrete counterpart
\begin{equation}
\argmin_{\substack{u = u^1,\ldots, u^{n}\\\boldsymbol{v}=\boldsymbol{v}^1,\ldots\boldsymbol{v}^{n-1}}} \sum_{i=1}^{n} \frac{1}{2}\left\|A^iu^i-f^i\right\|_2^2 + \alpha \left\|\nabla u^i\right\|_{1,2} + \sum_{i=1}^{n-1}\gamma \left\|u^i_t+\nabla u^i\cdot\boldsymbol{v}^i\right\|_1 + \beta\sum_{j=1}^2\left\|\nabla v^{i,j}\right\|_{1,2},
\label{oldJointModelTimeDiscrete}
\end{equation}
was proposed. For both, image sequence and velocity field, the respective total variation is used as a regularizer and the classical optical flow formulation from \eqref{opticalFlowKlassisch} connects image sequence and velocity field. From the perspective of image reconstruction the optical flow constraint acts as an additional temporal regularizer along the calculated motion fields $\boldsymbol{v}$. In \cite{burger2016variational} the existence of a minimizer for model \eqref{oldJointModel} under certain regularity assumptions for $\boldsymbol{v}$ and $\nabla\boldsymbol{v}$ has been shown. Despite the existence of a minimizer, calculating it is numerically challenging. Problems arise from the non-convexity and non-linearity of the optical flow term, from the non-differentiability of the L$^1$--norm and finally due to several linear operators acting on $u$ and $\boldsymbol{v}$. \\
The main drawback in terms of practical applications for \eqref{oldJointModelTimeDiscrete} is the restriction to displacements of small magnitude, which cannot be expected in practical applications. By exchanging the classical optical flow constraint with the beneficial linearization from \eqref{opticalFlowBesser} we create a model capable of handling large-scale displacements. The joint large-scale motion estimation and image reconstruction model then reads
\begin{equation}
	\argmin_{\substack{u = u^1,\ldots, u^{n}\\\boldsymbol{v}=\boldsymbol{v}^1,\ldots\boldsymbol{v}^{n-1}}} \sum_{i=1}^{n} \frac{1}{2}\|A^iu^i-f^i \|_2^2 + \alpha \left\|\nabla u^i\right\|_{1,2} + \sum_{i=1}^{n-1} \| \rho(\boldsymbol{v}^i,u^{i},u^{i+1}) \|_1 + \beta\sum_{j=1}^2\left\|\nabla v^{i,j}\right\|_{1,2}.
	\label{jointLargeScaleModel}
\end{equation}
By setting $\boldsymbol{\tilde{v}}^i=0$ in the equation $\rho(\boldsymbol{v}^i,u^{i},u^{i+1})$ we obtain the original energy.

\section{Numerical Scheme and Implementation}
We propose, referring to \cite{dirks}, a minimization scheme which alternatingly fixes $u$ and $\boldsymbol{v}$ and minimizes the energy for the other variable. The corresponding problems for static $u^1_k,\ldots,u^n_k$ in equation \eqref{subproblemV} resp. $\boldsymbol{v}_{k+1}^1,\ldots\boldsymbol{v}_{k+1}^{n-1}$ in equation \eqref{subproblemU} read
\begin{align}
\boldsymbol{v}_{k+1} &= \argmin_{\boldsymbol{v}=\boldsymbol{v}^1,\ldots\boldsymbol{v}^{n-1}} \sum_{i=1}^{n-1} \| \rho(\boldsymbol{v}^i,u^{i}_{k},u^{i+1}_{k}) \|_1 + \beta\sum_{j=1}^2\left\|\nabla v^{i,j}\right\|_{1,2}, \label{subproblemV} \\
u_{k+1} &= \argmin_{u = u^1,\ldots, u^{n}} \sum_{i=1}^{n} \frac{1}{2}\|A^iu^i-f^i \|_2^2 + \alpha \| \nabla{u}^{i}\|_1 + \sum_{i=1}^{n-1} \| \rho(\boldsymbol{v}^i_{k+1},u^{i},u^{i+1}) \|_1,
\label{subproblemU}
\end{align}
where the letter $k$ shall denote the iteration number. To start with the minimization problem in \eqref{subproblemV}, let us denote that in this formulation the evaluation of $u^{i+1}_k$ on intermediate grid points $x+\boldsymbol{\tilde{v}}$ is required. These evaluations can be generated by an interpolation scheme (linear, cubic etc.). Moreover, the linearization of the brightness constancy assumption requires a flow field $\boldsymbol{\tilde{v}}$ close to $\boldsymbol{v}$. This can be generated by an iterative coarse-to-fine approach, which solves the problem  on subsampled versions of $u^i$ first and using the upscaled result as initial value for the next finer version. Moreover, on each level several so-called warpings are performed, where the problem is solved for some initial $\boldsymbol{\tilde{v}}$ and the solution is used in the next step (see Section \ref{coarseToFinePyramid} and \cite{wedel2009improved,zach2007duality} for details). \\
Treating the minimization problem in equation \eqref{subproblemU} next, this problem incorporates the term 
\begin{align}
\rho(\boldsymbol{v}^i_{k+1},u^{i},u^{i+1}) = (\boldsymbol{v}^i_{k+1}-\boldsymbol{\tilde{v}}^i_{k+1})\cdot  \nabla\tilde{u}^{i+1} + \tilde{u}^{i+1} - u^i
\label{rhoSubU}
\end{align}
Let us recall that in the minimization process for $\boldsymbol{v}^i$, a warping scheme is applied, which creates subsequent versions of $\boldsymbol{v}$ and $\boldsymbol{\tilde{v}}$ which can be assumed to converge to a static quantity. Consequently, the difference $\boldsymbol{v}^i_{k+1}-\boldsymbol{\tilde{v}}^i_{k+1}$ becomes arbitrarily small, so it can be neglected in \eqref{rhoSubU} and the term simplifies to $\|\tilde{u}^{i+1} - u^i(x)\|_1$. We want to underline that for evaluating $\tilde{u}^{i+1}$ the same interpolation scheme as in the $\boldsymbol{v}$-problem has to be used to ensure numerical consistency. 
\subsection{Problem in $\boldsymbol{v}$}
The optical flow problem has no time-correspondence in $\boldsymbol{v}$ and thus reduces to $(n-1)$ subproblems of the form
\begin{align}
\argmin_{\boldsymbol{v}} \| \boldsymbol{v}\cdot  \nabla\tilde{u} + \tilde{u}_t \|_1 + \beta\sum_{j=1}^2\left\|\nabla v^{j}\right\|_{1,2},
\end{align}
where $\tilde{u}_t := -\boldsymbol{\tilde{v}}^i\cdot\nabla\tilde{u}^{i+1} + \tilde{u}^{i+1} - u^i$ and $\tilde{u} := \tilde{u}^{i+1}$. Applying a primal-dual algorithm \cite{pock2009algorithm,chambolle2011first} yields the following iterative scheme:
\begin{align*}
\boldsymbol{y}_1^{k+1} &= \frac{\boldsymbol{y}_1^k + \sigma^v_1 \nabla \bar{v}^k_1}{\max(1,\frac{\|\boldsymbol{y}_1^k + \sigma^v_1 \nabla \bar{v}^k_1\|_2}{\alpha})},\\
\boldsymbol{y}_2^{k+1} &=  \frac{\boldsymbol{y}_2^k + \sigma^v_2 \nabla \bar{v}^k_2}{\max(1,\frac{\|\boldsymbol{y}_2^k + \sigma^v_2 \nabla \bar{v}^k_2\|_2}{\alpha})},\\
y_3^{k+1} &= \max(-1,\min(1,y_3^k + \sigma^v_3 (\nabla \tilde{u} \cdot \bar{\boldsymbol{v}}^k + \tilde{u}_t))),\\
v_1^{k+1} &= v_1^k - \tau^v_1(\nabla^T\cdot\boldsymbol{y}_1^{k+1} + \tilde{u}_xy_3^{k+1}),\\
v_2^{k+1} &= v_2^k - \tau^v_2(\nabla^T\cdot\boldsymbol{y}_2^{k+1} + \tilde{u}_yy_3^{k+1}),\\
\bar{\boldsymbol{v}}^{k+1} &= 2\boldsymbol{v}^{k+1} - \boldsymbol{v}^{k},\\
\end{align*}
where vector-valued dual variables $\boldsymbol{y}_1,\boldsymbol{y}_2$, a regular dual variable $y_3$ and step sizes $\sigma^v_1,\sigma^v_2,\sigma^v_3,\tau^v_1$ and $\tau^v_2$ were introduced.
As a convergence criterion we use the primal-dual residual $r^{pd}_v$ from \cite{goldstein2013adaptive}. Applied to this problem, the residual can be calculated as
\begin{align*}
	p_1 &= \|\frac{v_1^{k} - v_1^{k+1}}{\tau^v_1} - \nabla^T\cdot(\boldsymbol{y}_1^{k+1}-\boldsymbol{y}_1^{k}) - \tilde{u}_x(y_3^{k+1}-y_3^{k})\|_1,\\
	p_2 &= \|\frac{v_2^{k} - v_2^{k+1}}{\tau^v_2} - \nabla^T\cdot(\boldsymbol{y}_2^{k+1}-\boldsymbol{y}_2^{k}) - \tilde{u}_y(y_3^{k+1}-y_3^{k})\|_1,\\
	d_1 &= \|\frac{\boldsymbol{y}_1^{k} - \boldsymbol{y}_1^{k+1}|}{\sigma_1} - \nabla (v_1^{k} - v_1^{k+1})\|_1,\\
	d_2 &= \|\frac{\boldsymbol{y}_2^{k} - \boldsymbol{y}_2^{k+1}|}{\sigma_2} - \nabla (v_2^{k} - v_2^{k+1})\|_1,\\
	d_3 &= \|\frac{y_3^{k} - y_3^{k+1}|}{\sigma_3} - \nabla \tilde{u} \cdot (\boldsymbol{v}^{k} - \boldsymbol{v}^{k+1})\|_1,\\
	r^{pd}_v &= p_1 + p_2 + d_1 + d_2 + d_3.
\end{align*}

\subsubsection{Coarse-to-fine pyramid}
\label{coarseToFinePyramid}
The optical flow calculation is incorporated into a coarse-to-fine pyramid with intermediate warping steps. This is done by creating a set with $nScales$ versions of the input images with a downsampling factor $\eta\in (0,1)$. To create these images, the input image is convolved with a Gaussian kernel with standard variation $\sigma_d$ and then extrapolated using a bicubic interpolation scheme.\\
The optical flow is calculated on the coarsest scale with initial $\tilde{\boldsymbol{v}}=0$ first. The result $\boldsymbol{v}$ is filtered using a two-dimensional median-filter of size $sizeMed$. Afterwards, a bicubic interpolation scheme is taken to calculate both $\tilde{\boldsymbol{v}}$ and $\nabla\tilde{\boldsymbol{v}}$. Then we solve the problem again on the same scale. This procedure is repeated $nWarps$ times, until the result $\boldsymbol{v}$ is upscaled to the next finer level using bicubic interpolation. To improve the convergence speed, it is beneficial to also upscale the dual variables $\boldsymbol{y}_1,\boldsymbol{y}_2$ and $y_3$.

\subsection{Problem u}
The problem in $u$ does not simplify to a series of time-independent problems, since individual frames are correlated by the flow. Consequently, the problem has to be solved in the whole space/time domain. First, we want to deduce that \eqref{subproblemU} can be rewritten in the form
\begin{align}
\argmin_{u} \frac{1}{2}\|\mathcal{A}u-f\|_2^2 + \alpha \| \bar{\nabla}u\|_{1,2} + \gamma\|\mathcal{W}u\|_1,\label{subproblemUmatrix}
\end{align}
where the identities $u=(u^1,\ldots,u^n),f=(f^1,\ldots,f^n),\mathcal{A}=diag(A^1,\ldots,A^n),\bar{\nabla} = diag(\nabla,\ldots,\nabla)$ hold obviously. For the last term, note that the optical flow term from Equation \eqref{subproblemU} consists of $(n-1)$ parts of the form $\tilde{u}^{i+1} - u^i$, where $\tilde{u}^{i+1}$ is a shifted version of $u^{i+1}$ by a velocity field $\boldsymbol{v}^i$ that can be evaluated using an interpolation scheme. This quantity can be approximated using a so-called interpolation or warping operator $W^{i+1}$ in which each row has weights at the columns according to the interpolation method (linear, cubic, spline etc.). Thus, we have $\tilde{u}^{i+1}\approx W^{i+1}u^{i+1}$ and the last part in $\eqref{subproblemU}$ reduces to $\|\mathcal{W}u\|_1$, where
\begin{align*}\mathcal{W} = \begin{bmatrix}
	-I  & W^2  & 0 & \ldots & \ldots     & 0  \\
	0  & -I &  W^3& 0 & \ldots & 0 \\
	\vdots & \ddots & \ddots & \ddots & \ddots & \vdots  \\
	\vdots	 & \ddots & 0 & -I & W^{n-2}&0\\
0	& \ldots & \ldots & 0 & -I & W^{n-1}
\end{bmatrix}.\end{align*}
Applying the primal-dual algorithm \cite{pock2009algorithm,chambolle2011first} to problem \eqref{subproblemUmatrix} we obtain the iterative scheme
\begin{align*}
y_1^{k+1} &= \frac{1}{\sigma^u_1 + 1}  (y_1^k + \sigma^u_1 (A\bar{u}^k - f)),\\
\boldsymbol{y}_2^{k+1} &= \frac{\boldsymbol{y}_2^k + \sigma^u_2 \bar{\nabla} \bar{u}^k}{\max(1,\frac{\|\boldsymbol{y}_2^k + \sigma^u_2 \bar{\nabla} \bar{u}^k\|_2}{\alpha})},\\
y_3^{k+1} &= \max(-\gamma,\min(\gamma,y_3^k + \sigma^u_3 \mathcal{W} \bar{u}^k )),\\
u^{k+1} &= u^k - \tau^u(A^Ty_1^{k+1} + \bar{\nabla}^T\cdot\boldsymbol{y}_2^{k+1} + \mathcal{W}^Ty_3^{k+1}),\\
\bar{u}^{k+1} &= 2u^{k+1} - u^{k},\\
\end{align*}
where, similar to the problem for $\boldsymbol{v}$, regular dual variables $y_1,y_3$, a vector-valued dual variable $\boldsymbol{y}_2$ and step-sizes $\sigma^u_1,\sigma^u_2,\sigma^u_3$ and $\tau^u$ were introduced. Once again, we use the primal-dual residual \cite{goldstein2013adaptive} as a convergence criterion that can be calculated as follows:
\begin{align*}
p &= \|\frac{u^{k} - u^{k+1}}{\tau} - \mathcal{A}^T(y_1^{k+1}-y_1^{k}) - \bar{\nabla}^T\cdot(\boldsymbol{y}_2^{k+1}-\boldsymbol{y}_2^{k}) - \mathcal{W}^T(y_3^{k+1}-y_3^{k})\|_1\\
d_1 &= \|\frac{{y}_1^{k} - {y}_1^{k+1}|}{\sigma} - A (u^{k} - u^{k+1})\|_1\\
d_2 &= \|\frac{\boldsymbol{y}_2^{k} - \boldsymbol{y}_2^{k+1}|}{\sigma} - \bar{\nabla} (u^{k} - u^{k+1})\|_1\\
d_3 &= \|\frac{y_3^{k} - y_3^{k+1}|}{\sigma} - \mathcal{W} ({u}^{k} - {u}^{k+1})\|_1\\
r^{pd}_u &= p + d_1 + d_2 + d_3
\end{align*}

\subsection{Initial Data}
Due to the fact that the initial formulation \eqref{jointLargeScaleModel} of the problem is highly non-linear, our alternating minimization scheme only solves the linear subproblems and will most likely end up in a local minimum. In our numerical evaluation, we found that initializing the image series $u$ with reasonable values we usually end up in a lower energy. This can for example be done by solving an ROF-problem for each image $u_i$, i.e. setting the weight $\gamma$ in  \eqref{subproblemUmatrix} to zero. \\
Another possibility to initialize the image sequence is to neglect the flow fields $\boldsymbol{v}$ in the whole problem.  One could for example expect a smooth time-derivative $u_t$ and solve to following modified problem
\begin{align}
\argmin_{u} \frac{1}{2}\|\mathcal{A}u-f\|_2^2 + \alpha \| \bar{\nabla}u\|_{1,2} + \frac{\epsilon}{2}\|u_t\|_2^2,\label{initialDataSmooth},
\end{align}
where $\epsilon$ is a small weight However, depending on the application, for large displacements this may introduce smearing effects and lead to worse results than the first method.

\subsection{Operator Discretization}
	We assume the underlying space-time grid to consist of the following set of discrete points:
	\begin{align*} 
	\left\lbrace (i,j,t) : i=0,\ldots,n_x,j=0,\ldots,n_y,t=0,\ldots,n_t \right\rbrace
	\end{align*}
	For the sake of simplicity we write down the discrete operators applied to an example variable $u$ respectively its adjoint $\boldsymbol{y}$. The required generalization to the appearing primal and dual variables is left to the reader. The discrete gradient is calculated using forward differences and Neumann boundary conditions. The corresponding adjoint operator consists of backward differences with Dirichlet boundary conditions. The resulting scheme reads:
	\begin{align*}
	u_x(i,j) &= \begin{cases} u(i+1,j)-u(i,j) &\mbox{if } i<n_x \\ 
	0 & \mbox{if } i=n_x \end{cases}\\
	u_y(i,j) &= \begin{cases} u(i,j+1)-u(i,j) &\mbox{if } j<n_y \\ 
	0 & \mbox{if } j=n_y \end{cases}\\
	\nabla\cdot \boldsymbol{y}(i,j) &= 
	\begin{cases} 
	y_1(i,j)-y_1(i-1,j) &\mbox{if } i>0 \\ 
	y_1(i,j) & \mbox{if } i=0\\
	-y_1(i-1,j) & \mbox{if } i=n_x 
	\end{cases}\\
	&+ 
	\begin{cases} 
	y_2(i,j)-y_2(i,j-1) &\mbox{if } j>0 \\ 
	y_2(i,j) & \mbox{if } j=0 \\
	-y_2(i,j-1) & \mbox{if } j=n_y .
	\end{cases}
	\end{align*} 
	\subsubsection{Warping Operator}
	The structure of the warping operator $W^i$ depends on the interpolation scheme used to evaluate $\tilde{u}(i,j) = u(i+v_1(i,j),j+v_2(i,j))$. In our observation, a cubic interpolation scheme creates by far the best results.\\
	The cubic interpolation scheme incorporates a 16-point neighborhood of the target position. This is given by all pairwise combinations $p_{kl} = (i_k,j_l)^T$ of
	\begin{align*}
		i_0 &= floor (i+v_1(i,j))-1, i_1 = i_0 + 1, i_2 = i_1 + 1, i_3 = i_2 + 1,\\
		j_0 &= floor (j+v_2(i,j))-1, j_1 = i_0 + 1, j_2 = j_1 + 1, j_3 = j_2 + 1.
	\end{align*} 
	Let us further denote 
	\begin{align*}
		\mathcal{I}_{1D}(p_0,p_1,p_2,p_3,x) :=& (-\frac{1}{2}p_0 + \frac{3}{2}p_1-\frac{3}{2}p_2 + \frac{1}{2}p_3)x^3 + (p_0-\frac{5}{2}p_1+2p_2-\frac{1}{2}p_3)x^2 \\
		&+ (-\frac{1}{2}p_0 + \frac{1}{2}p_2)x + p_1
	\end{align*}
	as the one-dimensional cubic interpolation at position $x$ using values at position $p_0,\ldots,p_3$. Then, we obtain the bicubic interpolation of $\tilde{u}(i,j)$ as a combination of two one-dimensional problems:
	\begin{align*}
		\tilde{p}_0 &:= \mathcal{I}_{1D}(p_{00},p_{01},p_{02},p_{03},j+v_2(i,j)),\\
		\tilde{p}_1 &:= \mathcal{I}_{1D}(p_{10},p_{11},p_{12},p_{13},j+v_2(i,j)),\\
		\tilde{p}_2 &:= \mathcal{I}_{1D}(p_{20},p_{21},p_{22},p_{23},j+v_2(i,j)),\\
		\tilde{p}_3 &:= \mathcal{I}_{1D}(p_{30},p_{31},p_{32},p_{33},j+v_2(i,j)),\\
		u(i+v_1(i,j),j+v_2(i,j)) &\approx \mathcal{I}_{1D}(\tilde{p}_0,\tilde{p}_1,\tilde{p}_2,\tilde{p}_3,i+v_1(i,j))
	\end{align*}
	Using this, one can build up the matrix $W^i$. For the row corresponding to the pair $(i,j)$ the quantities $v_1(i,j)$ and $v_2(i,j)$ are static, thus we can calculate $p_{kl}$. Plugging this into the interpolation formula and regrouping it with respect to $p_{kl}$ leads to an equation similar to $\sum_{k,l} \alpha_{kl} p_{kl}$. The weight $\alpha_{kl}$ is written at the column index corresponding to $p_{kl}$. In the boundary region we have to ensure that $i_0,j_0\geq 0, i_3\leq n_x$ and $j_3\leq n_y$. If one of these conditions is violated, the row is set to zero. The adjoint operator to $W^i$  is then simply the transposed matrix.

\section{Parameters}
\label{sec:parameters}
	In this section we give a short explanation of parameters and default choices for them, that are used in our experimental setup in Section \ref{section:experiments}. For parameters concerning the flow estimation we follow \cite{sun2014quantitative}, where various factors in the optical flow calculation are analyzed and a robust baseline approach is deduced.\\
	Stable stepsizes for the algorithm can be calculated using \cite{pock2011diagonal}. We recommend using diagonal matrices $\sigma$ and $\tau$ instead of scalar values for better convergence speed. Thus, the respective parameter choices refer to each element.
	\begin{itemize}
		\item $nSales$ The number of scales in the coarse-to-fine pyramid is dynamically chosen, such that the image size on the coarsest scale is 10 pixels minimum in each direction.
		\item $\eta$ The downsampling factor between subsequent levels is set to 0.8.
		\item $nWarps$ We do three warps per level of the pyramid.
		\item $sizeMed$ The median filtering is done in a 5x5 neighborhood.
		\item $\epsilon_v, \epsilon_u$ The stopping tolerance for both, the $u$ and the $\boldsymbol{v}$ problem is set to $10^{-6}$. 
		\item $\epsilon_{main}$ The stopping criterion for the main algorithm is set to $10^{-5}$.
		\item $nRes$ The residual is only updated every 100 iterations, which is sufficient since the algorithm typically needs more that 1000 iterations and the calculation is computationally expensive.
		\item $\alpha$ Typically, we have for the regularization weight in the image sequence $\alpha\in[0.005,0.1]$ depending on the noise in the input data. We usually do an initial calculation with $\alpha=0.02$.
		\item $\beta$ Since the flow part incorporates $\beta$ and $\gamma$ we propose to only adjust the ratio $\frac{\beta}{\gamma}\in [0.01,0.1]$ and to start with a regularization weight for both flow components of $\frac{\beta}{\gamma} = 0.02$.
		\item $\gamma$ The choice for $\gamma$ regulates how strong the images are coupled by the flow component. In our experiments, $\gamma=1$ is the usual choice. 
		\item $\sigma,\tau$  We calculate for the $\boldsymbol{v}$-problem $\sigma^v_1=\sigma^v_2=\frac{1}{2}, \sigma^v_3 = \frac{1}{\|\nabla\tilde{u}\|_1}, \tau^v_1 = \frac{1}{4 + |\tilde{u}_x|}$ and $\tau^v_2 = \frac{1}{4 + |\tilde{u}_y|}$. Analogously, we get for the $u$-problem $\sigma^u_1 = \frac{1}{\|A\|_1}, \sigma^u_2=\frac{1}{2}, \sigma^u_3=\frac{1}{\|\mathcal{W}\|_1}$ and $\tau^u = \frac{1}{\|A^T\|_1 + 4 + \|\mathcal{W}\|_1}$, where $\|\cdot\|_1$ for the matrix refers to the corresponding row-vector.
		\item $iterMainMax$ The maximum number of iterations for the main algorithm is set to 10 throughout our experiments.
	\end{itemize}

\section{Algorithm}

The algorithm can be separated into three problems: 
\begin{itemize}
	\item A $\boldsymbol{v}$-part \eqref{algorithmV} that calculates the flow field between every two subsequent images.
	\item A $u$-part \eqref{algorithmU} that solves the reconstruction problem for the image sequence using the flow information.
	\item A main algorithm \eqref{algorithmMain} that creates initial values and alternatingly calls the other two functions.
\end{itemize} 
The main algorithm uses a stopping criterion based on the difference between consecutive values for $u$ and $\boldsymbol{v}$ or alternatively stops after a defined number of iteration $iterMainMax$.\\
The flow part requires solving a series of minimization problems and, consequently, is the time consuming part of the main algorithm. The structure of the primal-dual algorithm allows parallelization of algorithm \eqref{algorithmV} and \eqref{algorithmU} on CPU or GPU to improve the runtime. 

\subsection{FlexBox}
An easier approach to minimize the arising variational problems for $u$ and $\boldsymbol{v}$ is given by \textbf{FlexBox} \cite{dirks2016flexible}, a flexible primal-dual toolbox. The basic framework for this optimization toolbox is written in MATLAB, but the software comes with an optional C$^{++}$ and CUDA module to massively enhance the runtime. \textbf{FlexBox} allows to formulate the primal variational problem term-by-term via MATLAB with only a few lines of code and calculates the solution in the fastest possible way. Creation of dual variables, calculation of step-sizes and transfer to the C$^{++}$ or CUDA solver are fully automated. The full sourcecode can be downloaded from http://www.flexbox.im.\\

\begin{algorithm}
		\caption{Subproblem for flow field $\boldsymbol{v}$}
		\label{algorithmV}
		\begin{algorithmic}
			\Statex
			\Function{problemV}{$u_1,u_2,\alpha$}
			\State Init $\boldsymbol{y}_1,\boldsymbol{y}_2,y_3,\boldsymbol{v}$ on coarsest scale
			\For{$s\gets 1,nScales$}
				\State Upsample  $\boldsymbol{y}_1,\boldsymbol{y}_2,y_3,\boldsymbol{v}$ to current scale using bicubic interpolation
				\For{$w\gets 1,nWarps$}
					\Let{$\boldsymbol{\tilde{v}}$}{$\boldsymbol{v}$}
					\State Calculate $\tilde{u}_2(x+\tilde{\boldsymbol{v}})$ and $\nabla\tilde{u}_2(x+\tilde{\boldsymbol{v}})$ w.r.t. $s$
					\While{$\epsilon_v<r^{pd}_v$}
						\Let{$\boldsymbol{y}_1^{k+1}$}{$\frac{\boldsymbol{y}_1^k + \sigma \nabla \bar{v}^k_1}{\max(1,\frac{\|\boldsymbol{y}_1^k + \sigma \nabla \bar{v}^k_1\|_2}{\alpha})}$}
						\Let{$\boldsymbol{y}_2^{k+1}$}{$\frac{\boldsymbol{y}_2^k + \sigma \nabla \bar{v}^k_2}{\max(1,\frac{\|\boldsymbol{y}_2^k + \sigma \nabla \bar{v}^k_2\|_2}{\alpha})}$}
						\Let{$y_3^{k+1}$}{$\max(-1,\min(1,y_3^k + \sigma (\nabla \tilde{u} \cdot \bar{\boldsymbol{v}}^k + \tilde{u}_t)))$}
						\Let{$v_1^{k+1}$}{$v_1^k - \tau(\nabla^T\cdot\boldsymbol{y}_1^{k+1} + \tilde{u}_xy_3^{k+1})$}	
						\Let{$v_2^{k+1}$}{$v_2^k - \tau(\nabla^T\cdot\boldsymbol{y}_2^{k+1} + \tilde{u}_yy_3^{k+1})$}
						\Let{$\bar{\boldsymbol{v}}^{k+1}$}{$2\boldsymbol{v}^{k+1} - \boldsymbol{v}^{k}$}
						\State Update $r^{pd}_v$ every $nRes$ iterations
					\EndWhile
					\State Apply median filter to $\boldsymbol{v}$
				\EndFor
			\EndFor
			\State \Return{$\boldsymbol{v}$}
			\EndFunction
		\end{algorithmic}
	\end{algorithm}
	\begin{algorithm}
		\caption{Subproblem for image sequence $u$}
		\label{algorithmU}
		\begin{algorithmic}
			\Statex
			\Function{problemU}{$f,\boldsymbol{v},\alpha,\gamma$}
			\State Generate operators $\mathcal{A},\bar{\nabla},\mathcal{W}$
			\While{$\epsilon_u<r^{pd}_u$}
			\Let{$y_1^{k+1}$}{$\frac{1}{\sigma + 1}  (y_1^k + \sigma (A\bar{u}^k - f))$}
			\Let{$\boldsymbol{y}_2^{k+1}$}{$\frac{\boldsymbol{y}_2^k + \sigma \bar{\nabla} \bar{u}^k}{\max(1,\frac{\|\boldsymbol{y}_2^k + \sigma \bar{\nabla} \bar{u}^k\|_2}{\alpha})}$}
			\Let{$y_3^{k+1}$}{$\max(-\gamma,\min(\gamma,y_3^k + \sigma \mathcal{W} \bar{u}^k ))$}
			\Let{$u^{k+1}$}{$u^k - \tau(A^Ty_1^{k+1} + \bar{\nabla}^T\cdot\boldsymbol{y}_2^{k+1} + \mathcal{W}^Ty_3^{k+1})$}
			\Let{$\bar{u}^{k+1}$}{$2u^{k+1} - u^{k}$}
			\State Update $r^{pd}_u$ every $nRes$ iterations
			\EndWhile
			\State \Return{$u$}
			\EndFunction
		\end{algorithmic}
	\end{algorithm}

	\begin{algorithm}
		\caption{Main iteration}
		\label{algorithmMain}
		\begin{algorithmic}
			\Statex
			\Function{mainFunction}{$f,\alpha,\beta,\gamma$}
			\Let{$iteration$}{0}
			\Let{$\boldsymbol{v}$}{0}
			\Let{$u$}{problemU($f,\boldsymbol{v},\alpha,0$)}
			\While{$r_{main}>tol_{main}$ or $iteration < iterMainMax$}
				\Let{$u_{old}$}{$u$}
				\Let{$\boldsymbol{v}_{old}$}{$\boldsymbol{v}$}
				\For{$i\gets 1,nImages-1$}
					\Let{$v_i$}{problemV($u_i,u_{i+1},\beta,\gamma$)}
				\EndFor
				\Let{$u$}{problemU($f,\boldsymbol{v},\alpha,\gamma$)}
				\Let{$r_{main}$}{$\|u-u_{old}\| + \|\boldsymbol{v}-\boldsymbol{v}_{old}\|$}
				\Let{$iteration$}{$iteration + 1$}
			\EndWhile
			\State \Return{$(u,\boldsymbol{v})$}
			\EndFunction
		\end{algorithmic}
	\end{algorithm}
	
	\subsection{Time-Continuous Model} We want to make clear that the given algorithm can be used to minimize the original time-continuous model from \cite{dirks} with only a few modifications. By leaving out the coarse-to-fine pyramid and the intermediate warping steps in the flow problem $\boldsymbol{v}$ (i.e. setting $\tilde{\boldsymbol{v}}=0$ in the evaluation of $\tilde{u}_2$ and $\nabla\tilde{u}_2$) we solve a classical optical flow problem. The sub-problem for the image sequence $u$ can be adjusted by exchanging the warping operator $\mathcal{W}$ with a discretization of the equation $\nabla u_2 \cdot \boldsymbol{v} + u_2-u_1 = K (u_1,u_2)^t$ with  $K = [-Id,diag(v_1)\nabla_x + diag(v_2)\nabla_y + Id]$
	does directly apply the calculated small-scale flows and has not to be changed at all.\\
	The limitations from \cite{dirks} of course apply and the model can only be used from sequences with motion of very small magnitude.

\section{Experimental Results}
\label{section:experiments}
\begin{figure}
	\centering
	\subfloat{\includegraphics[width=.32\linewidth]{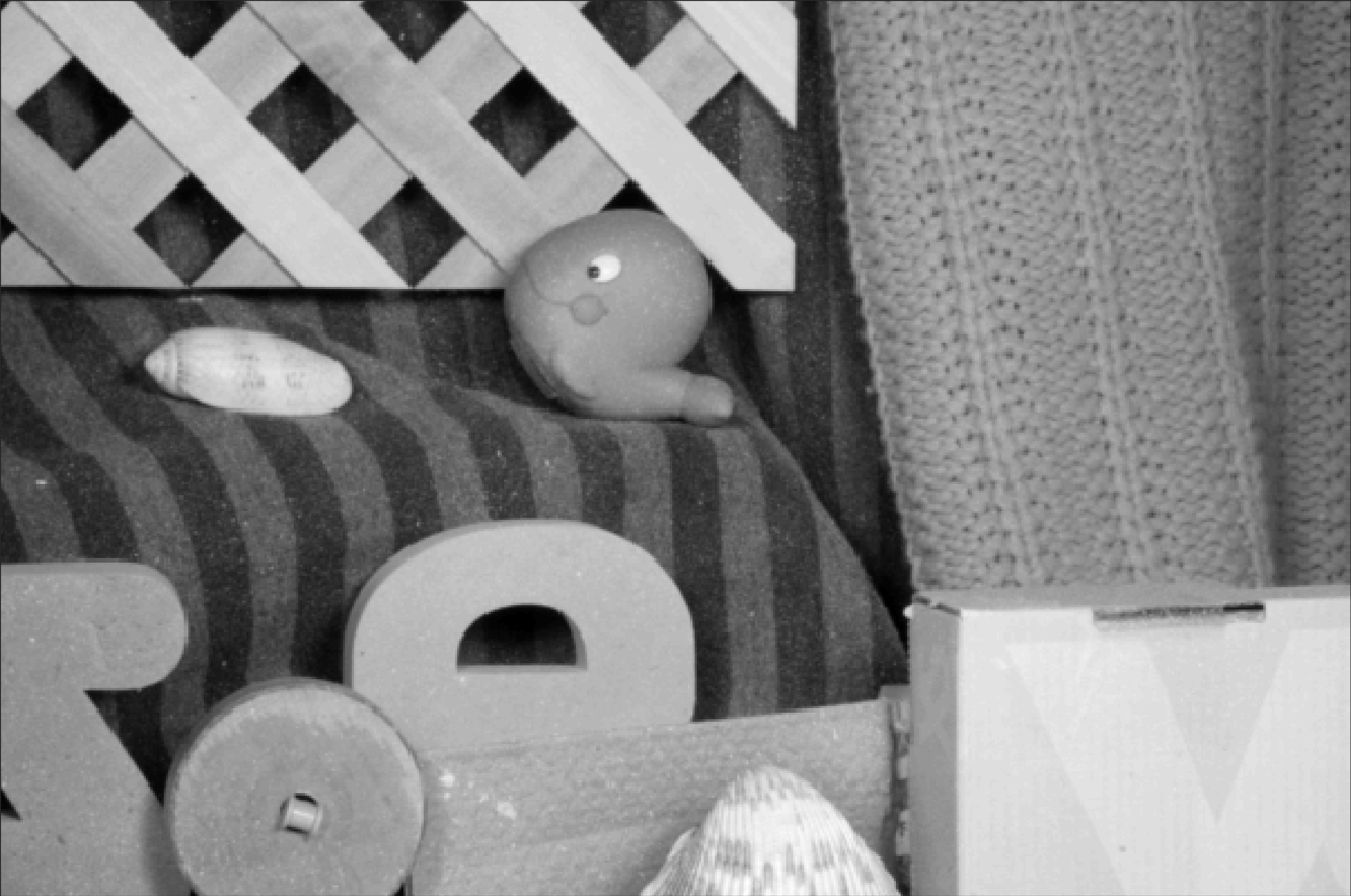}}\enskip
	\subfloat{\includegraphics[width=.32\linewidth]{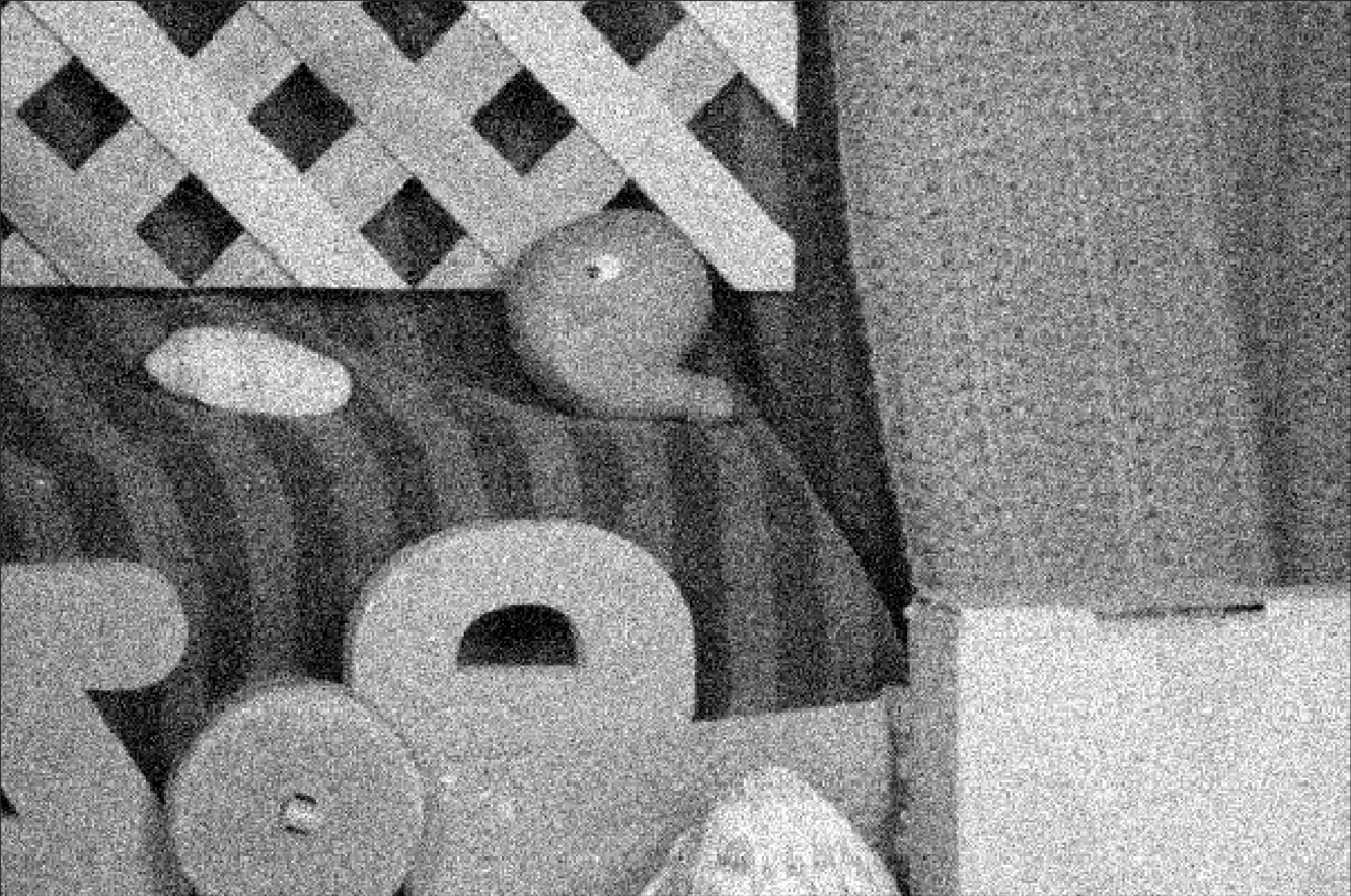}}\enskip
	\subfloat{\includegraphics[width=.32\linewidth]{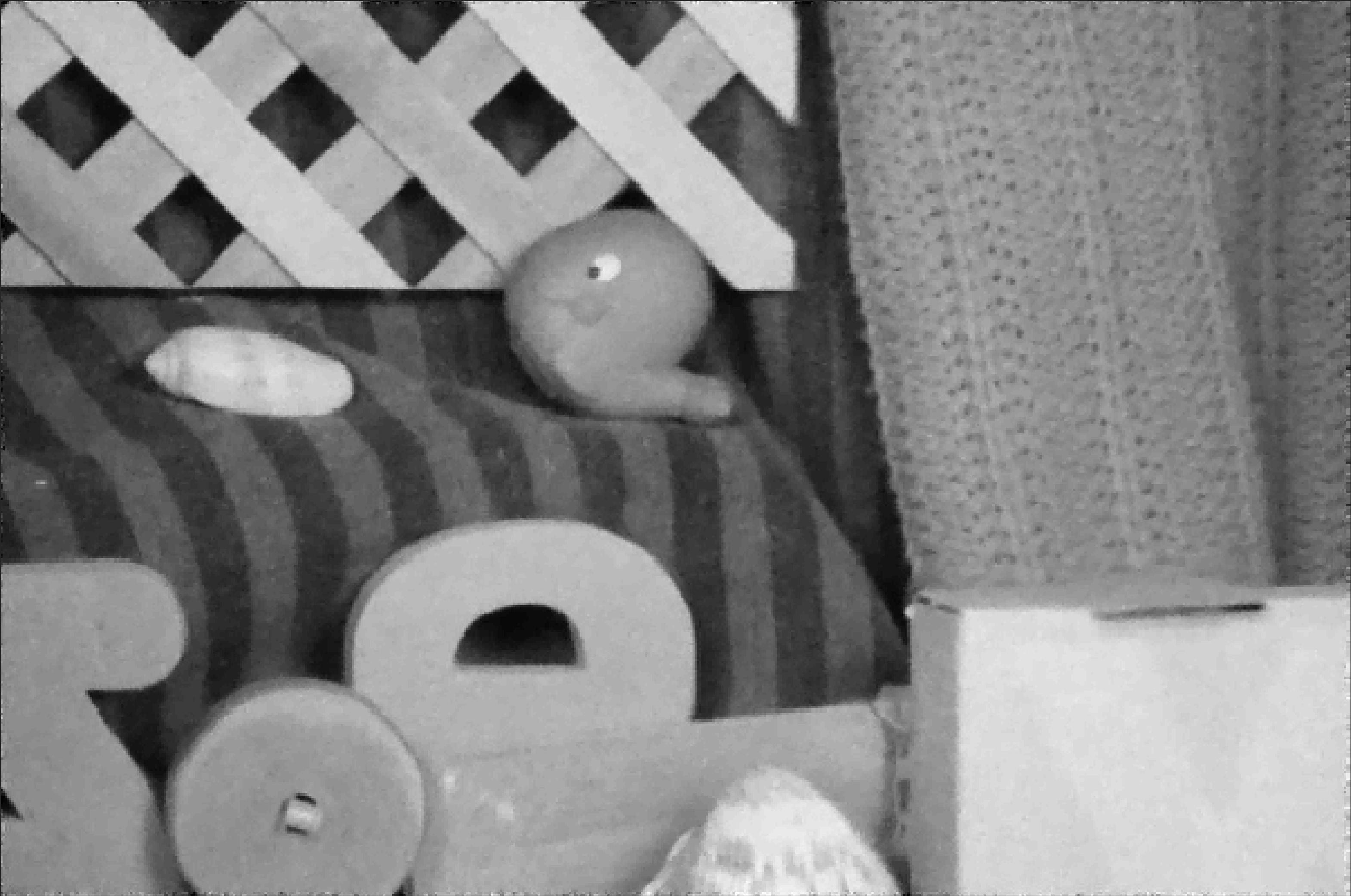}}\\
	\subfloat{\includegraphics[width=.32\linewidth]{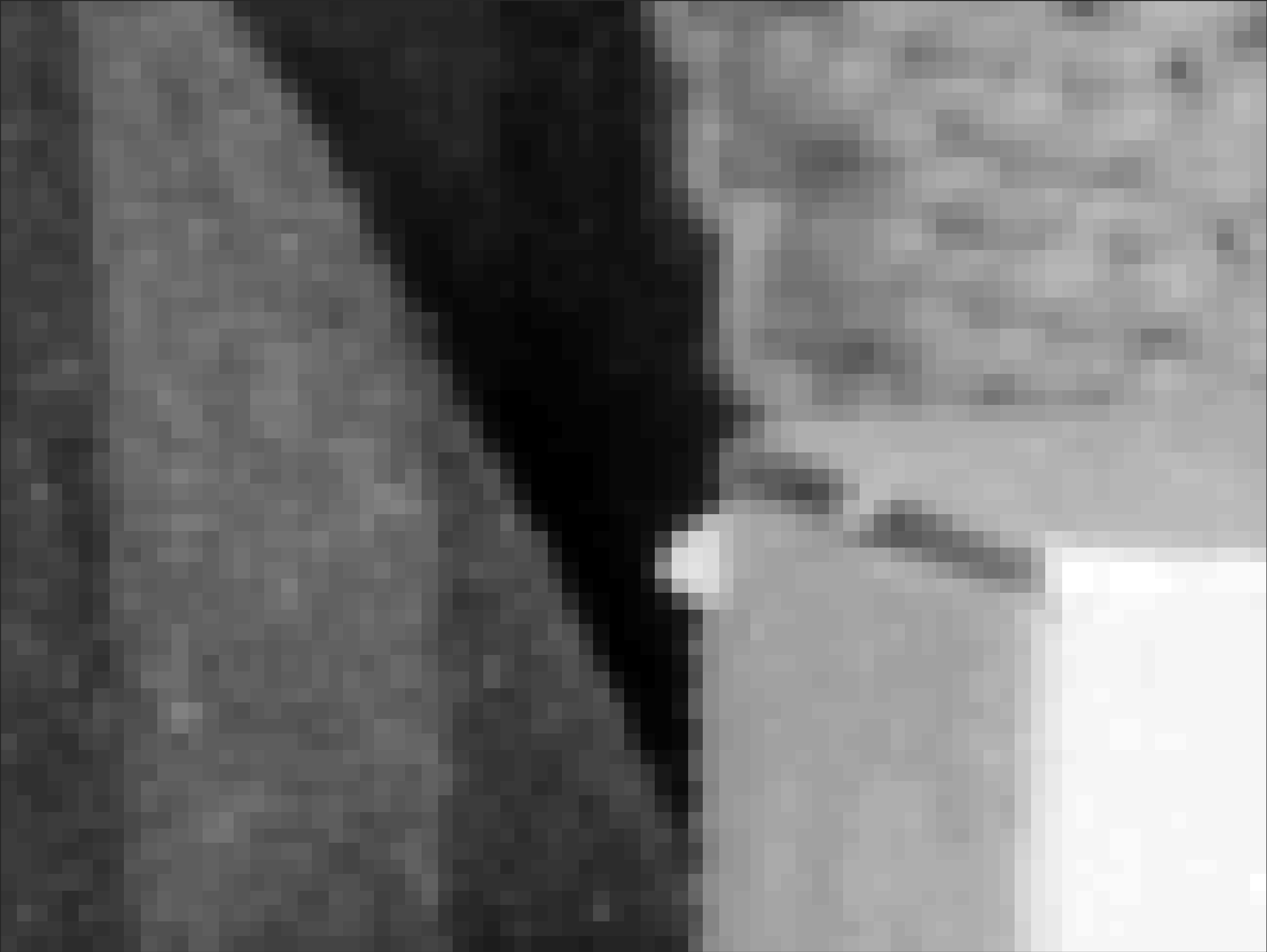}}\enskip
	\subfloat{\includegraphics[width=.32\linewidth]{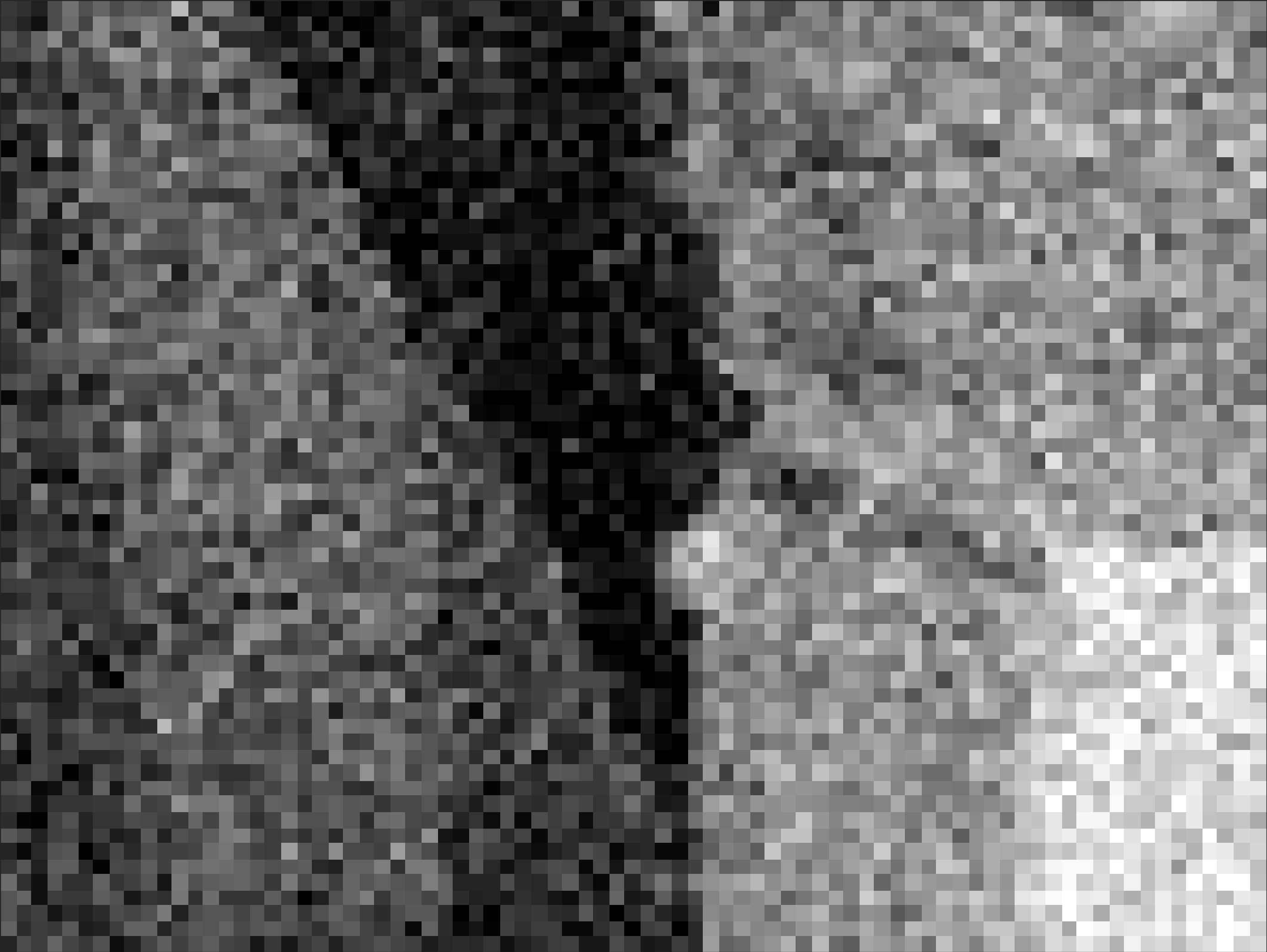}}\enskip
	\subfloat{\includegraphics[width=.32\linewidth]{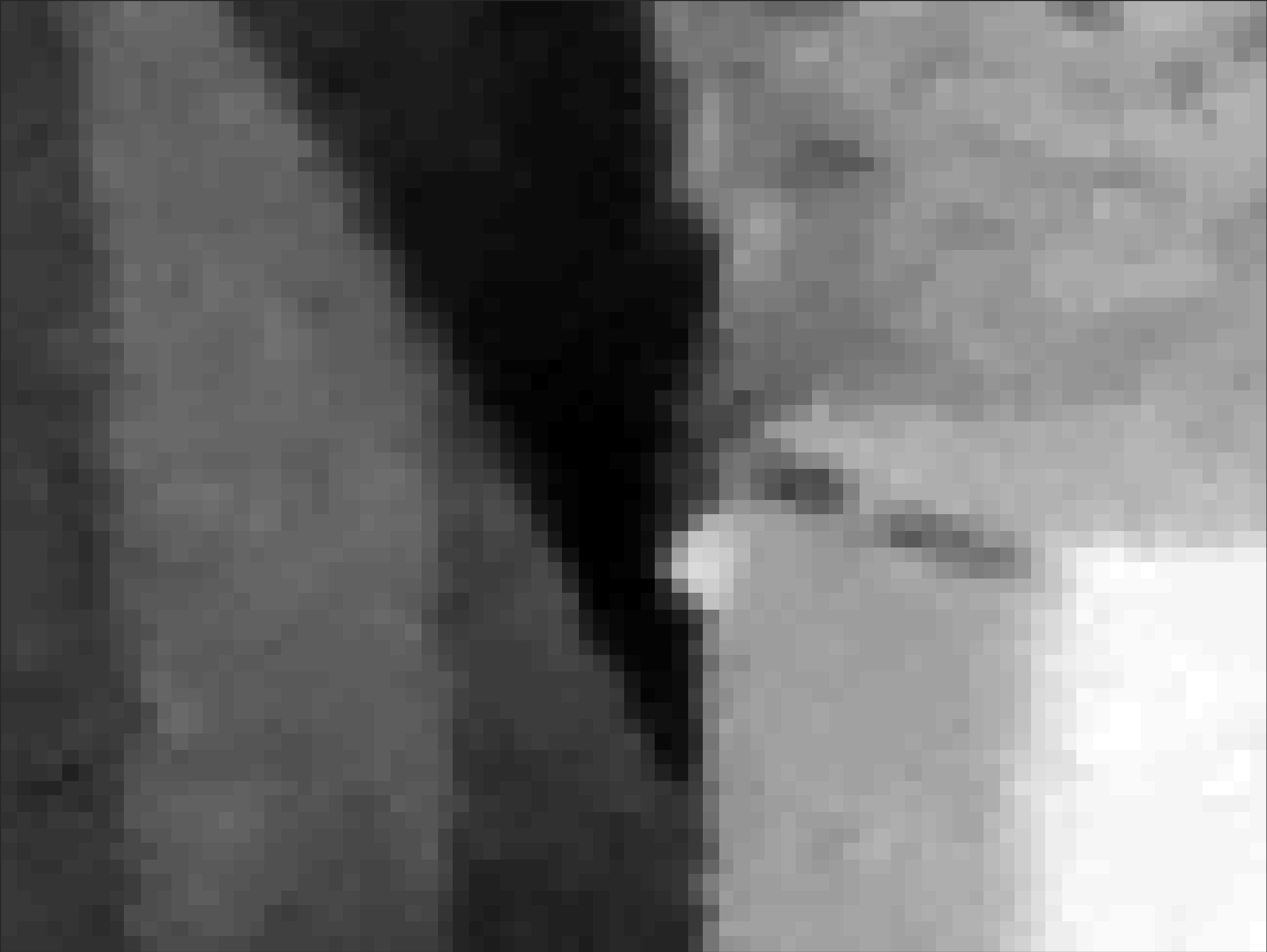}}\\
	\subfloat{\includegraphics[width=.48\linewidth]{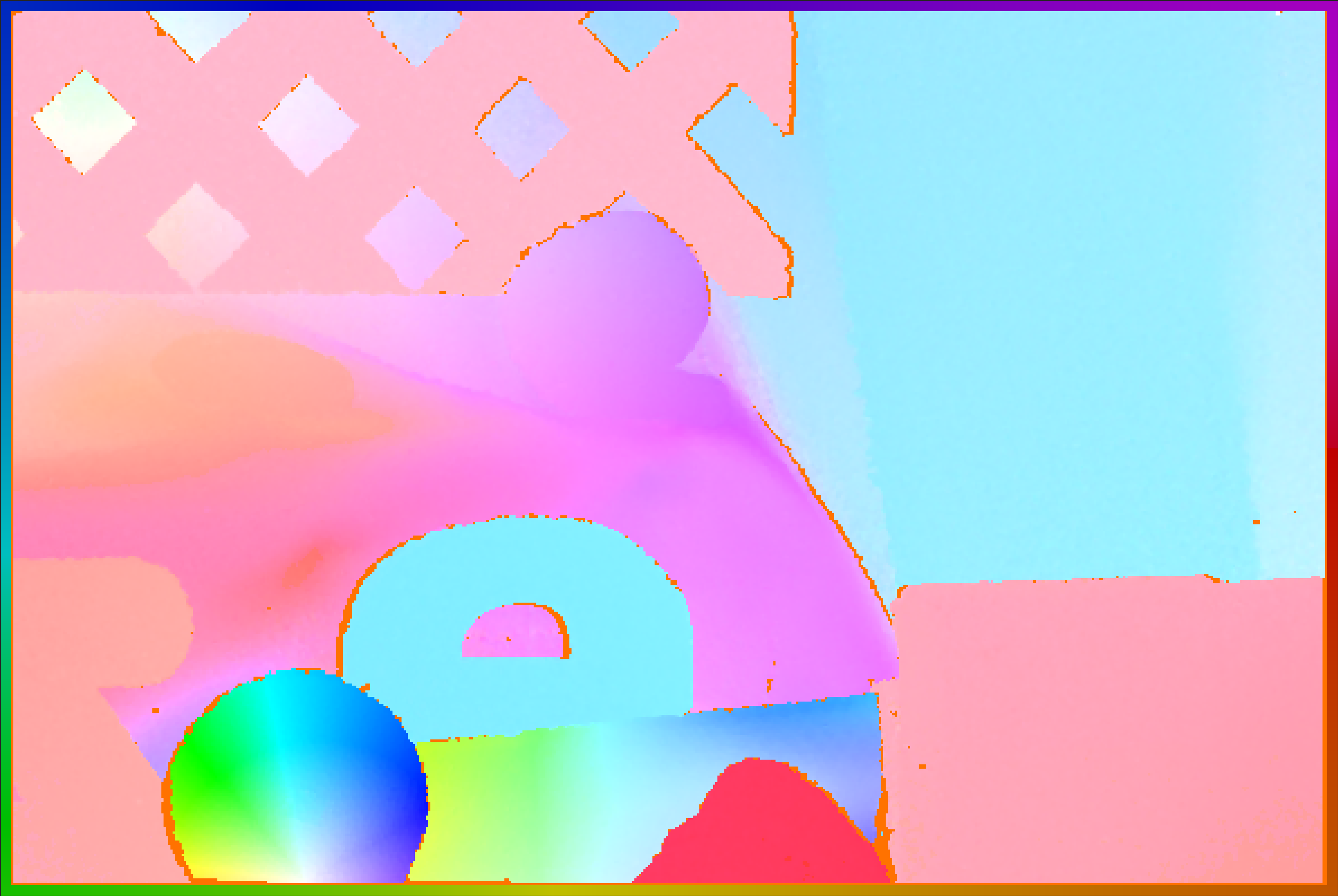}}\enskip
	\subfloat{\includegraphics[width=.48\linewidth]{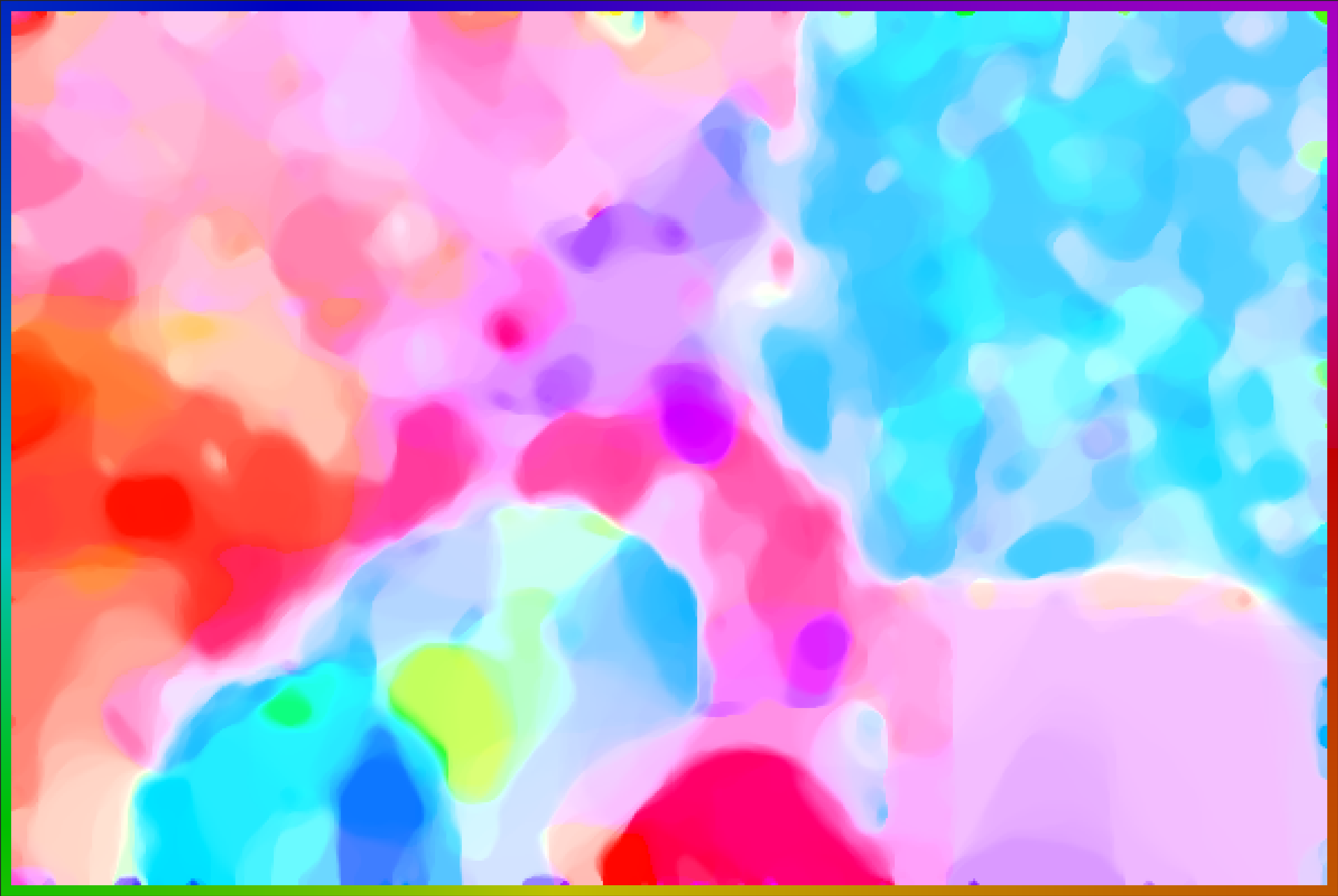}}\\
	\caption[]{Frame 4 of the Rubber Whale sequence from \cite{baker2011database}. The top row shows the full input image, disturbed counterpart and reconstruction. The middle row contains a zoom into the bottom right part of the image. In the bottom row we see ground truth and estimated flow field.}
	\label{fig:results}
\end{figure}

We evaluate the algorithm on grayscale datasets from the Middlebury optical flow database \cite{baker2011database}, which contain at least five consecutive frames and have available ground truth flow fields. Each image is disturbed with Gaussian noise ($\mu=0,\sigma^2=0.01$) to test both, the image denoising and the flow estimation capacities. We use static parameters presented in Section \ref{sec:parameters} throughout the evaluation.\\
To measure the image reconstruction quality, we calculate L$^2$-error, PSNR and SSIM (default MATLAB implementation) and average them with respect to the number of frames. To evaluate the calculated flow fields, we calculate endpoint error (EPE) \cite{otte1994optical} and angular error (AE) \cite{fleet1990computation,barron1994performance} between our result and the given ground-truth flow field. In the Middlebury sequence, the ground truth field is only given for the flow between the fourth and fifth frame.\\
The resulting errors are listed in Table \ref{tab:jointModelResultsOverview}. Figure \ref{fig:results} gives a visual overview of the denoising capabilities of our algorithm. Both, calculated errors and visual impression indicate a robust and superior reconstruction capability for dynamic image sequences. Moreover, the zoom in Figure \ref{fig:results} shows nicely recovered edges, well reconstructed structures and less staircasing effects than comparable total variation regularization techniques. The error in the flow field is relatively high, compared to similar motion estimation techniques. However, one has to take into account that usual benchmarks are calculated on clean input images. 

\begin{table}
	\centering
	\begin{tabular}{|c|c|l|l|l|l|l|}
		\hline
		&\textbf{Gove2} & \textbf{Grove3} & \textbf{Hydrangea} &  \textbf{RubberWhale} &  \textbf{Urban2}&  \textbf{Urban3}\\
		\hline
		SSIM & \GroveTwoSSIM & \GroveThreeSSIM & \HydrangeaSSIM & \RubberWhaleSSIM &\UrbanTwoSSIM & \UrbanThreeSSIM \\\hline
		L$^2$-Error & \GroveTwoLTwo & \GroveThreeLTwo & \HydrangeaLTwo & \RubberWhaleLTwo & \UrbanTwoLTwo & \UrbanThreeLTwo \\\hline
		PSNR & \GroveTwoSNR & \GroveThreeSNR & \HydrangeaSNR & \RubberWhaleSNR & \UrbanTwoSNR & \UrbanThreeSNR \\\hline
		EPE  & \GroveTwoEPE & \GroveThreeEPE & \HydrangeaEPE & \RubberWhaleEPE& \UrbanTwoEPE & \UrbanThreeEPE \\\hline
		AE & \GroveTwoAE & \GroveThreeAE & \HydrangeaAE & \RubberWhaleAE & \UrbanTwoAE & \UrbanThreeAE \\\hline
	\end{tabular}
	\caption{Image reconstruction and flow estimation errors for sequences (manually disturbed with Gaussian noise) from the Middlebury database \cite{baker2011database}.}
	\label{tab:jointModelResultsOverview}
\end{table}

\section{Conclusions}
In this work, we propose a detailed implementation of the work presented by Burger, Dirks and Sch\"onlieb \cite{burger2016variational} and extend their time-continuous model for small-scale flows to arbitrary large-scale motion between consecutive frames. The arising variational problem is minimized using a primal-dual method and the operator-discretization is explained in detail. We present concrete parameter choices and give a pseudo-code of the implemented algorithm. Finally, a short numerical evaluation based on data from the Middlebury optical flow database is presented.\\
An obvious possible extension would be to deal with color-images. The color-channels can be coupled here to improve the results. Moreover, an implementation to higher spatial dimensions is possible, which causes an increased runtime, and also may require a different motion model (e.g. mass-preservation) on the other hand.

\newpage
\bibliographystyle{siam}
\bibliography{citations}

\end{document}